\documentclass[11pt,a4paper]{article}
\usepackage[hyperref]{acl2020}
\usepackage{times}
\usepackage{latexsym}
\usepackage{amsfonts}
\usepackage{graphicx}
\usepackage{amsmath}
\usepackage{hyperref}
\usepackage{color}

\usepackage{microtype}

\aclfinalcopy % Uncomment this line for the final submission
\title{GCAN: Graph-aware Co-Attention Networks\\for Explainable Fake News Detection on Social Media}

\author{Yi-Ju Lu \\
  Department of Statistics \\ 
  National Cheng Kung University \\
  Tainan, Taiwan \\
  \texttt{l852888@gmail.com} \\\And
  Cheng-Te Li \\
  Institute of Data Science \\
  National Cheng Kung University \\
  Tainan, Taiwan \\
  \texttt{chengte@mail.ncku.edu.tw} \\}

\date{}

\begin{document}
\maketitle
\begin{abstract}
This paper solves the fake news detection problem under a more realistic scenario on social media. Given the source short-text tweet and the corresponding sequence of retweet users without text comments, we aim at predicting whether the source tweet is fake or not, and generating explanation by highlighting the evidences on suspicious retweeters and the words they concern. We develop a novel neural network-based model, Graph-aware Co-Attention Networks (GCAN), to achieve the goal. Extensive experiments conducted on real tweet datasets exhibit that GCAN can significantly outperform state-of-the-art methods by 16\% in accuracy on average. In addition, the case studies also show that GCAN can produce reasonable explanations. 
\end{abstract}

\section{Introduction}
\label{sec:intro}
Social media is indispensable in people's daily life, where users can express themselves, access news, and interact with each other. Information can further spread through the social network. Opinions and sentiments on source stories can be reflected by user participation and interaction. The convenient and low-cost essence of social networking brings collective intelligence, but at the same time leads to a negative by-product, the propagation of misinformation such as \textit{fake news}.

Fake news is a kind of news story possessing intentionally false information on social media~\citep{fakel17a,fake17sv2}. The widespread of fake news can mislead the public, and produce unjust political, economic, or psychological profit for some parties~\citep{fakel17b,fake17sv2}. Data mining and machine learning techniques were utilized to detect fake news~\citep{fake17sv1,cacm20}. Typical approaches rely on the content of new articles to extract textual features, such as n-gram and bag of words, and apply supervised learning (e.g., random forest and support vector machine) for binary classification~\citep{fake17sv1}. NLP researchers also learn advanced linguistic features, such as factive/assertive verbs and subjectivity~\citep{lang17} and writing styles and consistency~\citep{wstyle18}. Multi-modal context information is also investigated, such as user profiles~\citep{yangp12,crnn18p} and retweet propagation~\citep{csi17s,kshu19h}.

Nevertheless, there are still critical challenges in detecting fake news online. First, existing content-based approaches~\citep{textf11,wstyle18,kshu19h} require documents to be \textit{long} text, e.g., news articles, so that the representation of words and sentences can be better learned. However, tweets on social media are usually \textit{short} text~\citep{tshort15}, which produces severe data sparsity problem. Second, some state-of-the-art models~\citep{csi17s,crnn18p,kshu19h} require a rich collection of \textit{user comments} for every news story, to learn the opinions of retweeters, which usually provide strong evidences in identifying fake news. However, most users on social media tend to simply reshare the source story without leaving any comments~\citep{twitter10}. Third, some studies~\citep{trnn18} consider that the pathways of information cascade (i.e., retweets) in the social network are useful for classifying misinformation, and thus learn the representations of the tree-based propagation structures. However, it is costly to obtain the diffusion structure of retweets at most times due to privacy concerns~\citep{fpid18}. Many users choose to hide or delete the records of social interactions. Fourth, if the service providers or the government agencies desire to inspect who are the suspicious users who support the fake news, and which topics do they concern in producing fake news~\citep{fexp19}, existing models cannot provide explanations. Although dEFEND~\citep{kshu19h} can generate reasonable explanation, it requires both long text of source articles and text of user comments. 

This paper deals with fake news detection under a more realistic scenario on social media. We predict whether a source tweet story is fake, given only its \textit{short text} content and its \textit{retweet sequence of users}, along with \textit{user profiles}. That said, we detect fake news under three settings: (a) short-text source tweet, (b) no text of user comments, and (c) no network structures of social network and diffusion network. Moreover, we require the fake news detection model to be capable of \textit{explainability}, i.e., highlighting the evidence when determining a story is fake. The model is expected to point out the suspicious retweeters who support the spreading of fake news, and highlight the words they especially pay attention to from the source tweet. 

To achieve the goal, we propose a novel model, \underline{\textbf{G}}raph-aware \underline{\textbf{C}}o-\underline{\textbf{A}}ttention \underline{\textbf{N}}etwork (\textbf{GCAN})~\footnote{The Code of GCAN model is available and can be accessed via: \url{https://github.com/l852888/GCAN}}. We first extract user features from their profiles and social interactions, and learn word embeddings from the source short text. Then we use convolutional and recurrent neural networks to learn the \textit{representation of retweet propagation} based on user features. A graph is constructed to model the potential interactions between users, and the graph convolution network is used to learn the \textit{graph-aware representation of user interactions}. We develop a \textit{dual co-attention mechanism} to learn the correlation between the source tweet and retweet propagation, and the co-influence between the source tweet and user interaction. The binary prediction is generated based on the learned embeddings.

We summarize the contributions as follows. (1) We study a novel and more realistic scenario of fake news detection on social media. (2) For accurate detection, we develop a new model, GCAN, to better learn the representations of user interactions, retweet propagation, and their correlation with source short text. (3) Our dual co-attention mechanism can produce reasonable explanations. (4) Extensive experiments on real datasets demonstrate the promising performance of GCAN, comparing to state-of-the-art models. The GCAN explainability is also exhibited in case studies.

We organize this paper as follows. Section~\ref{sec:related} reviews the relevant approaches to fake news detection in social media. We describe the problem statement in Section~\ref{sec:prob}. Then in Section~\ref{sec:method}, the details of our proposed GCAN model will be elaborated. Section~\ref{sec:exp} demonstrates the evaluation settings and results. We conclude this work in Section~\ref{sec:conclude}.

\section{Related Work}
\label{sec:related}
% Existing studies on fake news detection can be divided into three approaches, content-based, user-based, and structure-based. Below we provide a brief review on such three aspects.

\textbf{Content-based} approaches rely on the text content to detect the truthfulness of news articles, which usually refer to long text. A variety of text characteristics are investigated for supervised learning, including TF-IDF and topic features~\citep{textf11}, language styles (e.g., part of speech, factive/assertive verbs, and subjectivity)~\citep{lang17}, writing styles and consistency~\citep{wstyle18}, and social emotions~\citep{emo19}. \citet{enquiry15} find the enquiry phrases from user responses are useful, and \citet{rnn16} use recurrent neural networks to learn better representations of user responses.
% so as to improve the performance. 

% \textbf{Content-based} approaches rely on the text content to detect the truthfulness of news articles, which usually refer to long documents. Typical text characteristics, including TF-IDF and topic features, are extracted for supervised learning~\citep{textf11}. Some work investigate the use of language styles, e.g., part of speech, factive/assertive verbs, and subjectivity~\citep{lang17}, to assess the truthfulness of news. The writing styles and consistency of news content with similar topics~\citep{wstyle18} are social emotions hidden in texts~\citep{emo19} are explored as well. Another direction is to further consider user response comments for a shared news story. \citet{enquiry15} find the enquiry phrases from user responses are useful, and \citet{rnn16} use recurrent neural networks to learn better representation of user responses so as to improve the performance. 

\textbf{User-based} approaches model the traits of users who retweet the source story. \citet{yangp12} extract account-based features, such as ``is verified'', gender, hometown, and number of followers. \citet{kshu19p} unveil user profiles between fake and real news are significantly different.
% , and prove that features distilled from user profiles contribute most in fake news detection. 
CRNN~\citep{crnn18p} devise a joint recurrent and convolutional network model (CRNN) to better represent retweeter's profiles.
Session-based heterogeneous graph embedding~\citep{shared18} is proposed to learn the traits of users so that they can be identified in shared accounts. However, since such a method relies on session information, it cannot be directly applied for fake news detection.
% To better represent profiles of users who share the source story, \citet{crnn18p} devise a joint recurrent and convolutional network model (CRNN) for the detection task. 
% Since the setting of CRNN is partially similar to this work, it is regarded as a strong competitor in our evaluation. 

\textbf{Structure-based} approaches leverage the propagation structure in the social network to detect fake news. 
% \citet{epi13s} first propose some simulation-based epidemiological models to characterize news cascades. 
\citet{lis16s} leverage the implicit information, i.e., hashtags and URLs, to connect conversations whose users do not have social links, and find such implicit info can improve the performance of rumor classification. \citet{kerl17s} create a kernel-based method that captures high-order patterns differentiating different types of rumors. \citet{trnn18} develop a tree-structured recursive neural networks to learn the embedding of rumor propagation structure. Although multi-relational graph embedding methods~\cite{marine19,osne19} are able to effectively learn how different types of entities (related to source news articles) interact with each other in a heterogeneous information network for classification tasks, they cannot be applied for the inductive setting, i.e., detecting the truthfulness of new-coming tweets.
% and make classification. 

\textbf{Hybrid-based} approaches consider and fuse multi-modal context information regarding the source tweets.
% , such as news textual content, attached images, user profiles, and social links, to identify misinformation.
CSI~\citep{csi17s} learns the sequential retweet features by incorporating response text and user profiles, and generates suspicious scores of users based on their social interactions. \citet{eann18h} develop an event adversarial neural network to learn transferable features by removing the event-specific features, along with convolutional neural networks to extract textual and visual features. dEFEND~\citep{kshu19h} jointly learns the sequential effect of response comments and the correlation between news content and comments, 
% leading to the state-of-the-art performance of fake news detection, 
and use an attention mechanism to provide explainability. 
% We will compare the proposed model with dEFEND.

We compare our work and the most relevant studies in Table~\ref{tab:related}. The uniqueness of our work lies in: targeting at short text, requiring no user response comments, and allow model explainability.

% \usepackage{graphicx}
\begin{table}[!t]
\centering
\caption{Comparison of related studies. Column notations: news story texts (NS), response comments (RC), user characteristics (UC), propagation structure (PS), social network (SN), and model explainability (ME). For the NS column, ``S'' and ``L'' indicates short and long text, respectively.}
\label{tab:related}
\resizebox{\linewidth}{!}{%
\begin{tabular}{ccccccc}
\hline
 & NS & RC & UC & PS & SN & ME \\ \hline\hline
~\citet{rnn16} & \checkmark (S) & \checkmark &  &  &  &  \\ \hline
~\citet{trnn18} & \checkmark (S) & \checkmark &  & \checkmark & \checkmark &  \\ \hline
~\citet{crnn18p} & \checkmark (S) &  & \checkmark & \checkmark &  &  \\ \hline
~\citet{csi17s} & \checkmark (S) & \checkmark & \checkmark &  &  &  \\ \hline
~\citet{kshu19h} & \checkmark (L) & \checkmark &  &  & \checkmark & \checkmark \\ \hline
Our work & \checkmark (S) &  & \checkmark & \checkmark & \checkmark & \checkmark \\ \hline
\end{tabular}%
}
\end{table}

\section{Problem Statement}
\label{sec:prob}
Let $\Psi=\{s_{1},s_{2}...s_{\left|\Psi\right|}\}$ be a set of tweet stories, and $U=\{u_{1},u_{2}...u_{\left|U\right|}\}$ be a set of users. Each $s_i\in \Psi$ is a short-text document (also called the \textit{source tweet}), given by $s_i=\{q^i_1,q^i_2,...,q^i_{l_i}\}$ indicating $l_i$ words in story $s_i$. Each $u_j\in U$ is associated with a user vector $\mathbf{x}_{j}\in \mathbb{R}^{d}$ representing the user feature with $d$ dimensions. When a news story $s_i$ is posted, some users will share $s_i$ and generate a sequence of retweet records, which is termed a \textit{propagation path}. Given a news story $s_i$, we denote its propagation path as $R_i=\{...,(u_j, \mathbf{x}_j,t_j),...\}$, where $(u_j, \mathbf{x}_j,t_j)$ depicts $j$-th user $u_j$ (with their feature vector $\mathbf{x}_j$) who retweets story $s_i$, and $j=1,2,...,K$ (i.e., $K=|R_i|$). We denote the set of users who retweet story $s_i$ as $U_i$. In $R_i$, we denote the user who originally shares $s_i$ as $u_1$ at time $t_1$. For $j>1$, user $u_j$ retweets $s_i$ at $t_j$ ($t_j>t_1$). Each story $s_i$ is associated with a binary label $y_i\in\{0,1\}$ to represent its truthfulness, where $y_i=0$ indicates story $s_i$ is true, and $y_i=1$ means $s_i$ is fake. 

Given a source tweet $s_i$, along with the corresponding propagation path $R_i$ containing users $u_j$ who retweet $s_i$ as well as their feature vectors $\mathbf{x}_j$, our goal is to predict the truthfulness $y_i$ of story $s_i$, i.e., binary classification. In addition, we require our model to highlight few users $u_j\in U_i$ who retweet $s_i$ and few words $q^i_k\in s_i$ that can interpret why $s_i$ is identified as a true or fake one. 

% Our problem settings in fake news detection are different from existing studies in terms of three aspects. First, our target is at short-text documents, whose language tends to be free, ambiguous, and even contain only very few words. Second, we rely on only retweet users without their comment texts. Such a setting is realistic as users in social media perform retweets more frequently than generating comments. Third, our model is further required to be explainable through highlighting some retweet users and some words that contributes most in detecting the truthfulness of the given story.

\section{The Proposed GCAN Model}
\label{sec:method}
We develop a novel model, Graph-aware Co-Attention Networks (GCAN), to predict fake news based on the source tweet and its propagation-based users. GCAN consists of five components. The first is \textit{user characteristics extraction}: creating features to quantify how a user participates in online social networking. The second is \textit{new story encoding}: generating the representation of words in the source tweet. The third is \textit{user propagation representation}: modeling and representing how the source tweet propagates by users using their extracted characteristics. The fourth is \textit{dual co-attention mechanisms}: capturing the correlation between the source tweet and users' interactions/propagation. The last is \textit{making prediction}: generating the detection outcome by concatenating all learned representations.
% Below we elaborate the details of each component.

\begin{figure}[!t]
\centering
\includegraphics[width=1.0\linewidth]{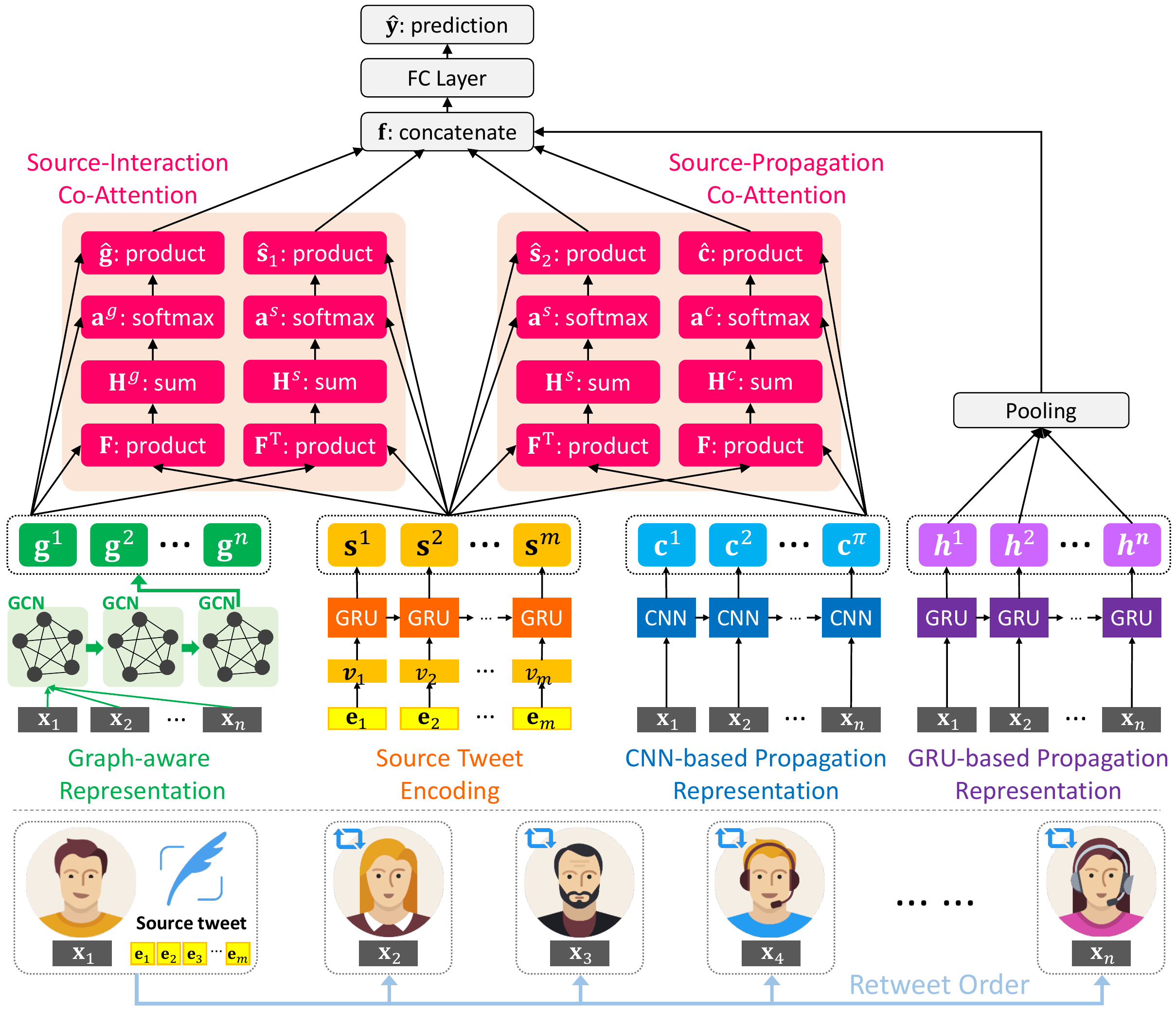}
\caption{The architecture of our GCAN model.}
\label{fig:model}
\end{figure}

\subsection{User Characteristics Extraction}
\label{sec:uce}
To depict how users participate in social networking, we employ their metadata and profiles to define the feature vector $\mathbf{x}_j$ of every user $u_j$. The extracted features are listed as follows: (1) number of words in a user's self-description, (2) number of words in $u_j$'s screen name, (3) number of users who follows $u_j$, (4) number of users that $u_j$ is following, (5) number of created stories for $u_j$, (6) time elapsed after $u_j$'s first story, (7) whether the $u_j$ account is verified or not, (8) whether $u_j$ allows the geo-spatial positioning, (9) time difference between the source tweet's post time and $u_j$'s retweet time, and (10) the length of retweet path between $u_j$ and the source tweet (1 if $u_j$ retweets the source tweet). Eventually, every user feature vector $\mathbf{x}_j\in \mathbb{R}^{v}$ is generated, where $v$ is the number of features.

\subsection{Source Tweet Encoding}
\label{sec:nse}
The given source tweet is represented by a word-level encoder. The input is the one-hot vector of each word in story $s_i$. Since the length of every source story is different, we perform zero padding here by setting a maximum length $m$. Let $\mathbf{E}=[e_1,e_2,...,e_m]\in \mathbb{R}^m$ be the input vector of source story, in which $e_m$ is the one-hot encoding of the $m$-th word. We create a fully-connected layer to generate word embeddings, $\mathbf{V}=[\mathbf{v}_1,\mathbf{v}_2,...,\mathbf{v}_m]\in \mathbb{R}^{d\times m}$, where $d$ is the dimensionality of word embeddings. The derivation of $\mathbf{V}$ is given by:
\begin{equation}
\mathbf{V}=\text{tanh}(\mathbf{W}_{w}\mathbf{E}+\mathbf{b}_w)
\end{equation}
where $\mathbf{W}_{w}$ is the matrix of learnable weights, and $\mathbf{b}_c$ is the bias term. Then, we utilize Gating Recurrent Units (GRU) \citep{chung14e} to learn the words sequence representation from $\mathbf{V}$. The source tweet representation learning can be depicted by: $\mathbf{s}_{t}=GRU(\mathbf{v}_{t})$, $t\in \{1,..., m\}$,
% \begin{equation}
% \begin{split}
% \mathbf{q}_{t}&=\sigma(\mathbf{U}_{q}\mathbf{v}_{t}+\mathbf{W}_{q}\mathbf{s}_{t-1} )\\
% \mathbf{k}_{t}&=\sigma(\mathbf{U}_{r}\mathbf{v}_{t}+\mathbf{W}_{k}\mathbf{s}_{t-1} )\\
% \tilde{\mathbf{s}}_{t}&=\text{tanh}(\mathbf{U}_{s}\mathbf{v}_{t}+\mathbf{s}_{t-1}\odot \mathbf{W}_{s}\mathbf{k}_{t})\\
% \mathbf{s}_{t}&=(1-\mathbf{q}_{t})\odot \mathbf{s}_{t-1}+\mathbf{q}_{t}\odot \tilde{\mathbf{s}}_{t}
% \end{split}
% \end{equation}
% where $\mathbf{U}_{q},\mathbf{U}_{k},\mathbf{U}_{s}\in \mathbb{R}^{d\times d}$ and $\mathbf{W}_{q},\mathbf{W}_{k},\mathbf{W}_{s}\in \mathbb{R}^{d\times d}$ are matrices of learnable parameters, 
where $m$ is the GRU dimensionality.
% $\sigma$ and $\text{tanh}$ are sigmoid and hyperbolic tangent activation functions, and $\odot$ is the element-wise vector multiplication operation, so 
We denote the source tweet representation as $\mathbf{S}=[\mathbf{s}^1,\mathbf{s}^2,...,\mathbf{s}^m]\in \mathbb{R}^{d\times m}$.

\subsection{User Propagation Representation}
\label{sec:ppr}
The propagation of source tweet $s_i$ is triggered by a sequence of users as time proceeds. We aim at exploiting the extracted user feature vectors $\mathbf{x}_j$, along with the user sequence spreading $s_i$, to learn user propagation representation. The underlying idea is that the user characteristics in real news propagations are different from those of fake ones. We make use of Gating Recurrent Units (GRU) and Convolutional Neural Network (CNN) to learn propagation representations.

Here the input is the sequence of feature vectors of users retweeting $s_i$, denoted by $PF(s_i)=\left \langle \mathbf{x}_1, \mathbf{x}_2,...,\mathbf{x}_t,...,\mathbf{x}_n \right \rangle$, where $n$ is the fixed length of observed retweets. If the number of users sharing $s_i$ is higher than $n$, we take the first $n$ users. If the number is lower than $n$, we resample users in $PF(s_i)$ until its length equals to $n$.

\textbf{GRU-based Representation.}
\label{gruemb}
Given the sequence of feature vectors $PF(s_i)=\left \langle ...,\mathbf{x}_t,..., \right \rangle$, we utilize GRU to learn the propagation representation. Each GRU state has two inputs, the current feature vector $\mathbf{x}_t$ and the previous state's output vector $\mathbf{h}_{t-1}$, and one output vector $\mathbf{h}_t$. The GRU-based representation learning can be depicted by:
% the following equations. 
$\mathbf{h}_{t}=GRU(\mathbf{x}_{t})$, $t\in \{1,..., n\}$, 
% \begin{equation}
% \begin{split}
% \mathbf{z}_{t}&=\sigma(\mathbf{U}_{z}\mathbf{x}_{t}+\mathbf{W}_{z}\mathbf{h}_{t-1} )\\
% \mathbf{r}_{t}&=\sigma(\mathbf{U}_{r}\mathbf{x}_{t}+\mathbf{W}_{r}\mathbf{h}_{t-1} )\\
% \tilde{\mathbf{h}}_{t}&=\text{tanh}(\mathbf{U}_{h}\mathbf{x}_{t}+\mathbf{h}_{t-1}\odot \mathbf{W}_{h}\mathbf{r}_{t})\\
% \mathbf{h}_{t}&=(1-\mathbf{z}_{t})\odot \mathbf{h}_{t-1}+\mathbf{z}_{t}\odot \tilde{\mathbf{h}}_{t}
% \end{split}
% \end{equation}
% where $\mathbf{U}_{z},\mathbf{U}_{r},\mathbf{U}_{h}\in \mathbb{R}^{d\times v}$ and $\mathbf{W}_{z},\mathbf{W}_{r},\mathbf{W}_{h}\in \mathbb{R}^{d\times d}$ are matrices of learnable parameters, 
where $n$ is the dimensionality of GRU.
% and $\sigma$ and $\text{tanh}$ are sigmoid and hyperbolic tangent activation functions
% , and $\odot$ is the element-wise vector multiplication operation. 
We generate the final GRU-based user propagation embedding $\mathbf{h} \in \mathbb{R}^{d}$ by average pooling, given by $\mathbf{h}=\frac{1}{n}\sum_{t=1}^{n}\mathbf{h}_t$.

\textbf{CNN-based Representation.}
\label{cnnemb}
We take advantage of 1-D convolution neural network to learn the sequential correlation of user features in $PF(s_i)$. We consider $\lambda$ consecutive users at one time to model their sequential correlation, i.e., $\left \langle \mathbf{x}_t,...,\mathbf{x}_{t+\lambda-1} \right \rangle$. Hence the filter is set as $\mathbf{W}_f\in \mathbb{R}^{\lambda \times v}$. Then the output representation vector $\mathbf{C}\in \mathbb{R}^{d\times (t+\lambda-1)}$ is given by
\begin{equation}
\mathbf{C}=\text{ReLU}(\mathbf{W}_{f}\cdot \mathbf{X}_{t:t+\lambda-1}+b_{f})
\end{equation}
where $\mathbf{W}_{f}$ is the matrix of learnable parameters, $ReLU$ is the activation function, $\mathbf{X}_{t:t+\lambda-1}$ depicts sub-matrices whose first row's index is from $t=1$ to $t=n-\lambda+1$, and $b_f$ is the bias term.

\subsection{Graph-aware Propagation Representation}
\label{sec:gpr}
We aim at creating a graph to model the potential interaction among users who retweet source story $s_i$. The idea is that some correlation between users with particular characteristics can reveal the possibility that the source tweet is fake. To fulfill such an idea, a graph $\mathcal{G}^i=(U_i,\mathcal{E}_i)$ is constructed for the set of users who share source story $s_i$ (i.e., $U_i$), where $\mathcal{E}_i$ is the corresponding edge set. Since the true interactions between users are unknown, we consider $\mathcal{G}^i$ is a fully-connected graph, i.e., $\forall e_{\alpha\beta}\in \mathcal{E}_i$, $u_{\alpha}\in U_i, u_{\beta}\in U_i$, and $u_{\alpha} \neq u_{\beta}$, $|\mathcal{E}_i|=\frac{n\times(n-1)}{2}$. 
To incorporate user features in the graph, each edge $e_{\alpha\beta}\in \mathcal{E}_i$ is associated with a weight $\omega_{\alpha\beta}$, and the weight is derived based on cosine similarity between user feature vectors $\mathbf{x}_{\alpha}$ and $\mathbf{x}_{\beta}$, given by $\omega_{\alpha\beta}=\frac{\mathbf{x}_{\alpha}\cdot\mathbf{x}_{\beta}}{\left \| \mathbf{x}_{\alpha} \right \| \left \| \mathbf{x}_{\beta} \right \|}$. We use matrix $\mathbf{A}=[\omega_{\alpha\beta}]\in \mathbb{R}^{n\times n}$ to represent weights between any pair of nodes $u_{\alpha}$ and $u_{\beta}$ in graph $\mathcal{G}^i$.

A graph convolution network (GCN) layer \citep{gcn16} is created based on the constructed graph $\mathcal{G}^i$ for source tweet $s_i$. A GCN is a multi-layer neural network that performs on graph data and generates embedding vectors of nodes according to their neighborhoods. GCN can capture information from a node's direct and indirect neighbors through stacking layer-wise convolution. Given the matrix $\mathbf{A}$ for graph $\mathcal{G}^i$, and $\mathbf{X}$ depicting the matrix of feature vectors for users in $\mathcal{G}^i$, the new $g$-dimensional node feature matrix $\mathbf{H}^{(l+1)}\in \mathbb{R}^{n\times g}$ can be derived by
\begin{equation}
\mathbf{H}^{(l+1)} = \rho(\tilde{\mathbf{A}}\mathbf{H}^{(l)}\mathbf{W}_l),
\end{equation}
where $l$ is the layer number, $\tilde{\mathbf{A}}=\mathbf{D}^{-\frac{1}{2}}\mathbf{A}\mathbf{D}^{-\frac{1}{2}}$ is the normalized symmetric weight matrix ($\mathbf{D}_{ii}=\sum_j \mathbf{A}_{ij}$), and $\mathbf{W}_l\in\mathbb{R}^{d\times g}$ is the matrix of learnable parameters at the $l$-th GCN layer. $\rho$ is an activation function, i.e., a ReLU $\rho(x)=\text{max}(0,x)$. Here $\mathbf{H}^{(0)}$ is set to be $\mathbf{X}$. We choose to stack two GCN layers in derive the learned graph-aware representation, denoted as $\mathbf{G}\in \mathbb{R}^{g\times n}$.

\subsection{Dual Co-attention Mechanism}
\label{sec:dcm}
We think the evidence of fake news can be unveiled through investigating which parts of the source story are concerned by which kinds of retweet users, and fake clues can be reflected by how retweet users interact with each other. Therefore, we develop a \textit{dual co-attention mechanism} to model the mutual influence between the source tweet (i.e., $\mathbf{S}=[\mathbf{s}^1,\mathbf{s}^2,...,\mathbf{s}^m]$) and user propagation embeddings (i.e., $\mathbf{C}=[\mathbf{c}^1,\mathbf{c}^2,...,\mathbf{c}^{n-\lambda+1}]$ from Section~\ref{cnnemb}), and between the source tweet and graph-aware interaction embeddings (i.e., $\mathbf{G}=[\mathbf{g}^1,\mathbf{g}^2,...,\mathbf{g}^n]$ from Section~\ref{sec:gpr}). Equipped with co-attention learning, our model is capable of the explainability by looking into the attention weights between retweet users in the propagation and words in the source tweet. In other words, by extending the co-attention formulation~\citep{coatt16}, the proposed dual co-attention mechanism aims to attend to the source-tweet words and graph-aware interaction users simultaneously (source-interaction co-attention), and also attend to the source-tweet words and propagated users simultaneously (source-propagation co-attention). 

\textbf{Source-Interaction Co-attention.}
We first compute a proximity matrix $\mathbf{F}\in \mathbb{R}^{m\times n}$ as: $\mathbf{F}=\text{tanh}(\mathbf{S}^\top\mathbf{W}_{sg}\mathbf{G})$, where $\mathbf{W}_{sg}$ is a $d\times g$ matrix of learnable parameters. By treating the proximity matrix as a feature, we can learn to predict source and interaction attention maps, given by
\begin{equation}
\begin{split}
\mathbf{H}^s&=\text{tanh}(\mathbf{W}_s\mathbf{S}+(\mathbf{W}_g\mathbf{G})\mathbf{F}^\top)\\
\mathbf{H}^g&=\text{tanh}(\mathbf{W}_g\mathbf{G}+(\mathbf{W}_s\mathbf{S})\mathbf{F})\\
\end{split}
\end{equation}
where $\mathbf{W}_s\in \mathbf{R}^{k\times d},\mathbf{W}_g\in \mathbf{R}^{k\times g}$ are matrices of learnable parameters. The proximity matrix $\mathbf{F}$ can be thought to transforming user-interaction attention space to source story word attention space, and vice versa for its transpose $\mathbf{F}^\top$. Then we can generate the attention weights of source words and interaction users through the softmax function:
\begin{equation}
\begin{split}
\mathbf{a}^s&=\text{softmax}(\mathbf{w}_{hs}^{\top}\mathbf{H}^s)\\
\mathbf{a}^g&=\text{softmax}(\mathbf{w}_{hg}^{\top}\mathbf{H}^g)
\end{split}
\end{equation}
where $\mathbf{a}^s\in \mathbb{R}^{1\times m}$ and $\mathbf{a}^g\in \mathbb{R}^{1\times n}$ are the vectors of attention probabilities for each word in the source story and each user in the interaction graph, respectively. $\mathbf{w}_{hs}, \mathbf{w}_{hg} \in \mathbb{R}^{1\times k}$ are learnable weights. Eventually we can generate the attention vectors of source story words and interaction users through weighted sum using the derived attention weights, given by
\begin{equation}
\hat{\mathbf{s}}_1=\sum_{i=1}^{m}\mathbf{a}^s_i\mathbf{s}^i~,~~~~\hat{\mathbf{g}}=\sum_{j=1}^{n}\mathbf{a}^g_j\mathbf{g}^j
\end{equation}
where $\hat{\mathbf{s}}_1\in \mathbb{R}^{1\times d}$ and $\hat{\mathbf{g}}\in \mathbb{R}^{1\times g}$ are the learned co-attention feature vectors that depict how words in the source tweet are attended by users who interact with one another.

\textbf{Source-Propagation Co-attention.} The process to generate the co-attention feature vectors, $\hat{\mathbf{s}}_2\in \mathbb{R}^{1\times d}$ and $\hat{\mathbf{c}}\in \mathbb{R}^{1\times d}$, for the source story and user propagation, respectively, is the same as source-interaction co-attention, i.e., creating another proximity matrix to transform them into each other's space. We skip the repeated details due to the page limit.

Note that the GRU-based user representations are not used to learn the interactions with the source tweet. The reason is that how user profiles in the retweet sequence look like is also important, as suggested by CRNN~\cite{crnn18p}, and should be emphasized separately. Nevertheless, the CNN-based user representations (i.e., features that depict the sequence of user profiles) has been used in the co-attention mechanism to learn their interactions with source tweet.

\subsection{Make Prediction}
\label{sec:mpd}
We aim at predicting fake news using the source-interaction co-attention feature vectors $\hat{\mathbf{s}}_1$ and $\hat{\mathbf{g}}$, the source-propagation feature vectors $\hat{\mathbf{s}}_2$ and $\hat{\mathbf{c}}$, and the sequential propagation feature vector $\mathbf{h}$. Let $\mathbf{f}=[\hat{\mathbf{s}}_1, \hat{\mathbf{g}}, \hat{\mathbf{s}}_2, \hat{\mathbf{c}}, \mathbf{h}]$ which is then fed into a multi-layer feedforward neural network that finally predicts the label. We generate the binary prediction vector $\hat{\mathbf{y}}=[\hat{y}_0,\hat{y}_1]$, where $\hat{y}_0$ and $\hat{y}_1$ indicate the predicted probabilities of label being $0$ and $1$, respectively. It can be derived through
\begin{equation}
% \begin{split}
\hat{\mathbf{y}}=\text{softmax}(\text{ReLU}(\mathbf{f}\mathbf{W}_f+\mathbf{b}_f)),
% \forall j\in\mathbf{[q]}\\
% \hat{\mathbf{y}}&=\text{softmax}(\mathbf{Z}_q)
% \end{split}
\end{equation}
where $\mathbf{W}_f$ is the matrix of learnable parameters, 
% $\mathbf{q}$ is the number of hidden layers, $\mathbf{Z}_j$ is the output of the $\mathbf{j}^{th}$ hidden layer 
and $\mathbf{b}_f$ is the bias term. The loss function is devised to minimize the cross-entropy value:
\begin{equation}
\mathcal{L}(\Theta)=-y\log(\hat{y}_1)-(1-y)\log(1-\hat{y}_0)
\end{equation}
where $\Theta$ denotes all learnable parameters in the entire neural network. We choose the Adam optimizer to learn $\Theta$ as it can determine the learning rate abortively.

\section{Experiments}
\label{sec:exp}
We conduct experiments to answer three questions: (1) whether our GCAN model is able to achieve satisfactory performance of fake news detection, compared to state-of-the-art methods? (2) how does each component of GCAN contribute to the performance? (3) can GCAN generate a convincing explanation that highlights why a tweet is fake?

% \usepackage{graphicx}
\begin{table}[!t]
\centering
\caption{Statistics of two Twitter datasets.}
\label{tab:data-stat}
% \resizebox{\linewidth}{!}{%
\begin{tabular}{lrr}
\hline
 & \textbf{Twitter15} & \textbf{Twitter16} \\ \hline
\# source tweets & 742 & 412 \\ %\hline
\# true & 372 & 205 \\ %\hline
\# fake & 370 & 207 \\ %\hline
\# users & 190,868 & 115,036 \\ %\hline
avg. retweets per story & 292.19 & 308.70\\ %\hline
avg. words per source & 13.25 & 12.81 \\ \hline
\end{tabular}%
% }
\end{table}

\begin{table*}[!t]
\centering
\caption{Main results. The best model and the best competitor are highlighted by \textbf{bold} and \underline{underline}, respectively.}
\label{tab:overallexp}
\begin{tabular}{l|cccc|cccc}
\hline
\multicolumn{1}{c}{}&\multicolumn{4}{c}{\textbf{Twitter15}}&\multicolumn{4}{c}{\textbf{Twitter16}}\\
\hline
Method&F1&Rec&Pre&Acc&F1&Rec&Pre&Acc\\\hline
DTC&0.4948&0.4806&0.4963&0.4949&0.5616&0.5369&0.5753&0.5612\\
SVM-TS&0.5190&0.5186&0.5195&0.5195&0.6915&0.6910&0.6928&0.6932\\
mGRU&0.5104&0.5148&0.5145&0.5547&0.5563&0.5618&0.5603&0.6612\\
RFC&0.4642&0.5302&0.5718&0.5385&0.6275&\underline{0.6587}&\underline{0.7315}&0.6620\\
tCNN&0.5140&0.5206&0.5199&0.5881&0.6200&0.6262&0.6248&0.7374\\
CRNN&0.5249&0.5305&0.5296&0.5919&\underline{0.6367}&0.6433&0.6419&\underline{0.7576}\\
CSI&\underline{0.7174}&\underline{0.6867}&\underline{0.6991}&0.6987&0.6304 &0.6309 &0.6321 &0.6612 \\
dEFEND&0.6541&0.6611&0.6584&\underline{0.7383}&0.6311&0.6384&0.6365&0.7016\\\hline
\textbf{GCAN-G}&0.7938&0.7990&0.7959&0.8636&0.6754&0.6802&0.6785&0.7939\\
\textbf{GCAN}& \textbf{0.8250}& \textbf{0.8295}& \textbf{0.8257}& \textbf{0.8767}& \textbf{0.7593}& \textbf{0.7632}& \textbf{0.7594}& \textbf{0.9084}\\\hline
\textbf{Improvement}& \textbf{15.0\%}& \textbf{20.8\%}& \textbf{18.1\%}& \textbf{18.7\%}& \textbf{19.3\%}& \textbf{15.9\%}& \textbf{3.8\%}& \textbf{19.9\%}\\\hline
\end{tabular}
\end{table*}

\subsection{Datasets and Evaluation Settings}
\textbf{Data.} Two well-known datasets compiled by~\citet{kerl17s}, Twitter15 and Twitter16, are utilized. Each dataset contains a collection of source tweets, along with their corresponding sequences of retweet users. We choose only ``true'' and ``fake'' labels as the ground truth. Since the original data does not contain user profiles, we use user IDs to crawl user information via Twitter API.

\textbf{Competing Methods.} We compare our GCAN with the state-of-the-art methods and some baselines, as listed below. (1) \textbf{DTC}~\citep{textf11}: a decision tree-based model combining user profiles and the source tweet. (2) \textbf{SVM-TS}~\citep{svmts15}: a linear support vector machine classifier that utilizes the source tweet and the sequence of retweet users' profiles. (3) \textbf{mGRU}~\citep{rnn16}: a modified gated recurrent unit model for rumor detection, which learns temporal patterns from retweet user profile, along with the source's features. (4) \textbf{RFC}~\citep{rfc17}: an extended random forest model combining features from retweet user profiles and the source tweet. (5) \textbf{CSI}~\citep{csi17s}: a state-of-the-art fake news detection model incorporating articles, and the group behavior of users who propagate fake news by using LSTM and calculating the user scores. (6) \textbf{tCNN}~\citep{tcnn18}: a modified convolution neural network that learns the local variations of user profile sequence, combining with the source tweet features. (7) \textbf{CRNN}~\citep{crnn18p}: a state-of-the-art joint CNN and RNN model that learns local and global variations of retweet user profiles, together with the resource tweet. (8) \textbf{dEFEND}~\citep{kshu19h}: a state-of-the-art co-attention-based fake news detection model that learns the correlation between the source article's sentences and user profiles.

\textbf{Model Configuration.} Our model is termed ``\textbf{GCAN}''. To examine the effectiveness of our graph-aware representation, we create another version ``\textbf{GCAN-G}'', denoting our model without the graph convolution part. For both our models and competing methods, we set the number of training epochs to be 50. The hyperparameter setting of GCAN is: number of retweet users = 40, word embedding dim = 32, GRU output dim = 32, 1-D CNN output filter size = 3, 1-D CNN output dim = 32, and GCN output dim = 32. The hyperparameters of competing methods are set by following the settings mentioned in respective studies.

\textbf{Metrics \& Settings.} The evaluation metrics include Accuracy, Precision, Recall, and F1. We randomly choose 70\% data for training and 30\% for testing. The conducted train-test is repeated 20 times, and the average values are reported.

\subsection{Experimental Results}
\textbf{Main Results.} The main results are shown in Table~\ref{tab:overallexp}. We can clearly find that the proposed GCAN significantly outperforms the best competing methods over all metrics across two datasets, improving the performance by around 17\% and 15\% on average in Twitter15 and Twitter16, respectively. Even without the proposed graph-aware representation, GCAN-G can improve the best competing method by 14\% and 3\% on average in Twitter15 and Twitter16, respectively. Such promising results prove the effectiveness of GCAN for fake news detection. The results also imply three insights. First, GCAN is better than GCAN-G by 3.5\% and 13\% improvement in Twitter15 and Twitter16, respectively. This exhibits the usefulness of graph-aware representation. Second, the dual co-attention mechanism in GCAN is quite powerful, as it clearly outperforms the best non-co-attention state-of-the-art model CSI. Third, while both GCAN-G and dEFEND are co-attention-based, additional sequential features learned from the retweet user sequence in GCAN-G can significantly boost the performance. 

\begin{figure}[!t]
\centering
\includegraphics[width=1.0\linewidth]{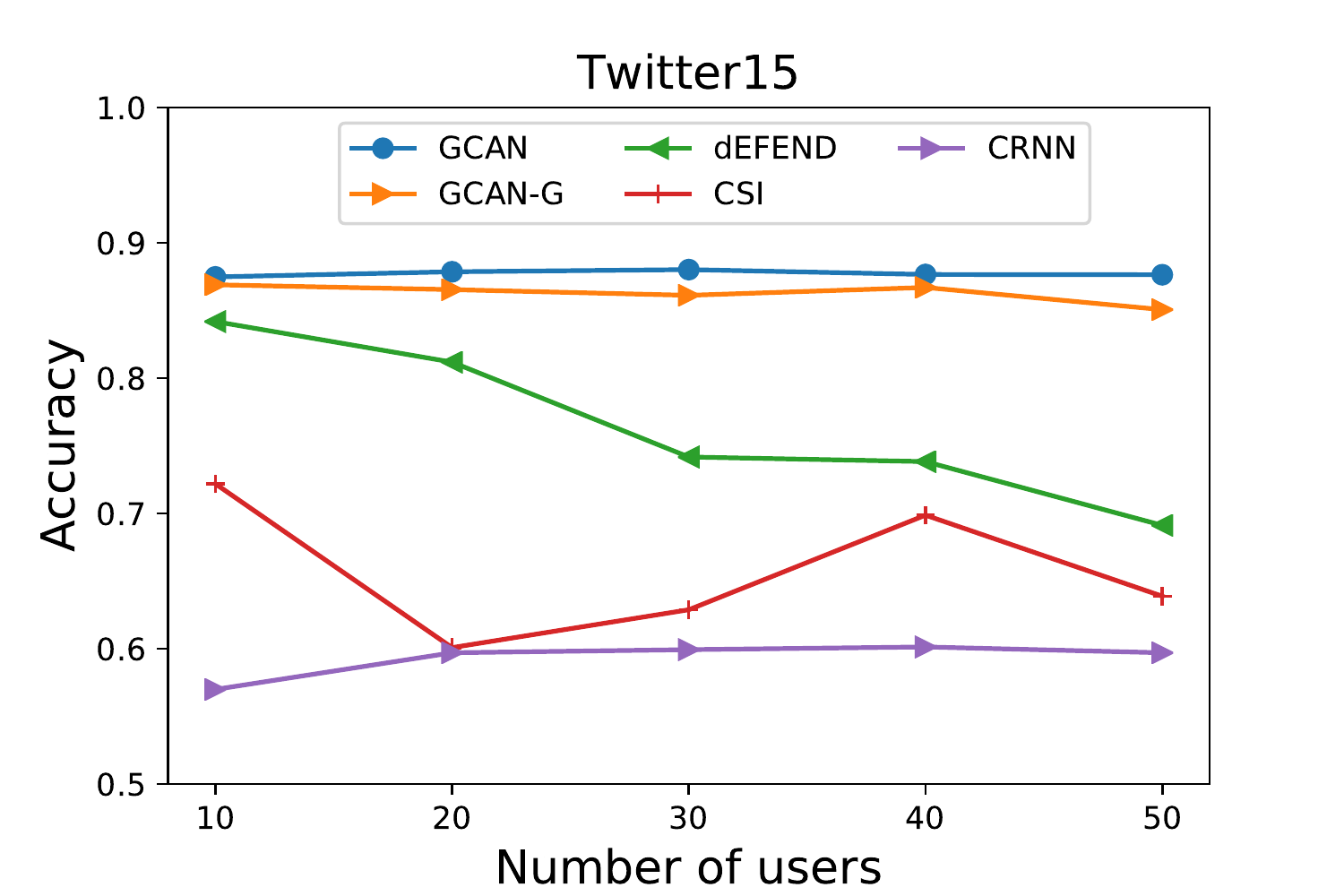}
\caption{Accuracy by \# retweet users in Twitter15.}\label{fig:compare15}
\end{figure}

\begin{figure}[!t]
\centering
\includegraphics[width=1.0\linewidth]{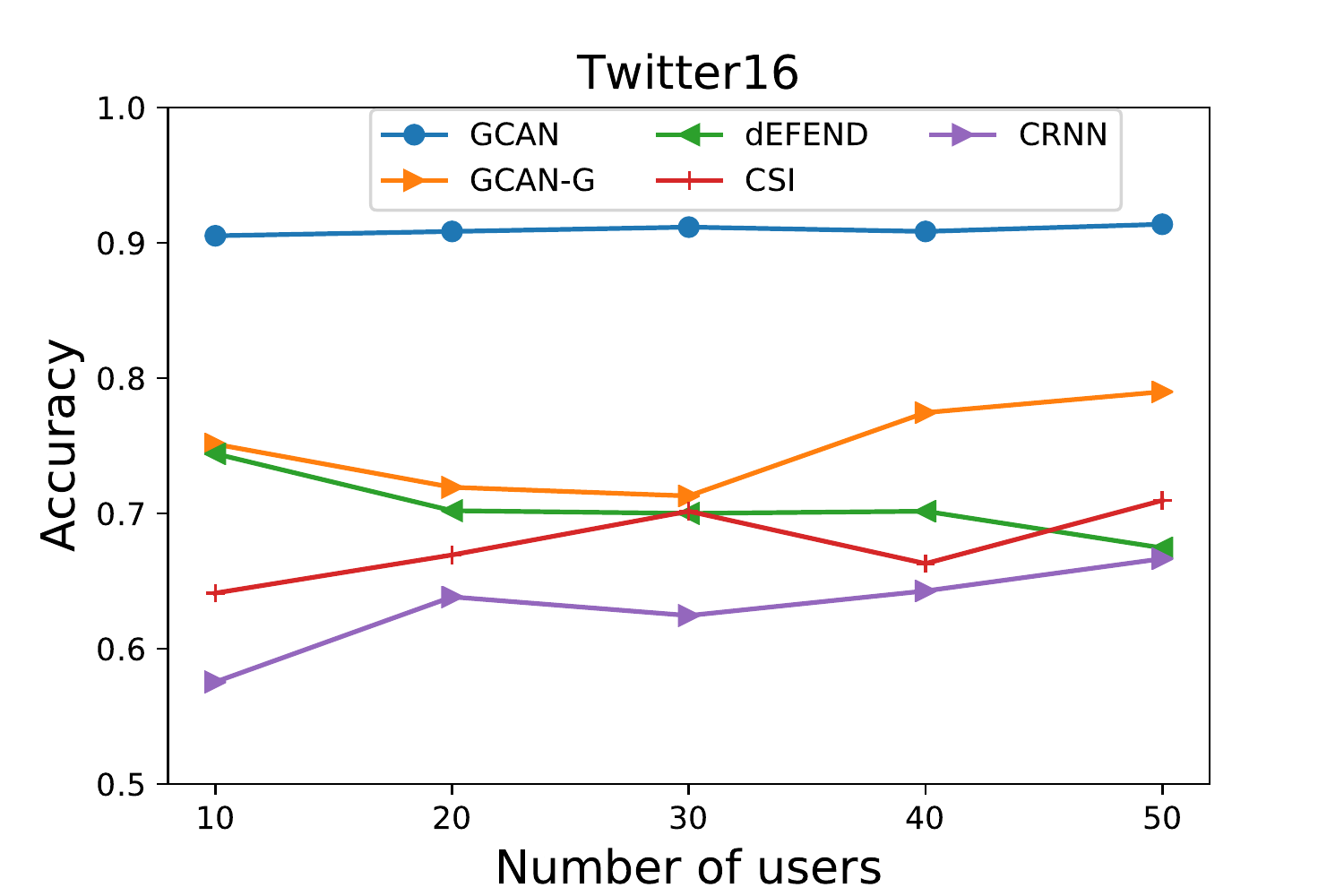}
\caption{Accuracy by \# retweet users in Twitter16.}\label{fig:compare16}
\end{figure}

\textbf{Early Detection.} We further report the performance (in only Accuracy due to page limit) by varying the number of observed retweet users per source story (from $10$ to $50$), as exhibited in Figure~\ref{fig:compare15} and Figure~\ref{fig:compare16}. It can be apparently found that our GCAN consistently and significantly outperforms the competitors. Even with only ten retweeters, GCAN can still achieve 90\% accuracy. Such results tell GCAN is able to generate accurate early detection of the spreading fake news, which is crucial when defending misinformation. 

\begin{figure}[!t]
\centering
\includegraphics[width=1.0\linewidth]{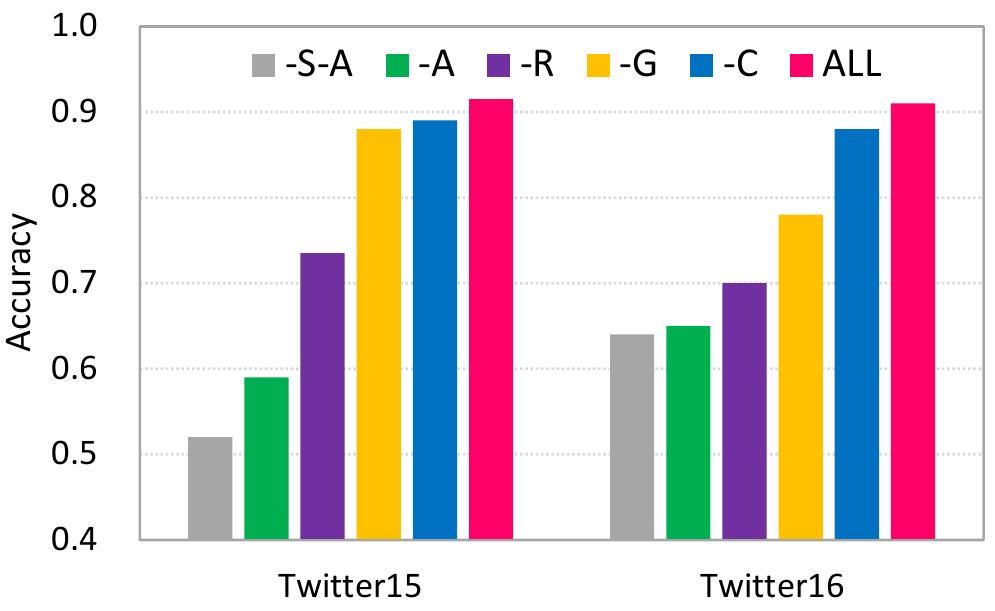}
\caption{GCAN ablation analysis in Accuracy.}
\label{fig:ablation}
\end{figure}

\textbf{Ablation Analysis.} We report how each of GCAN component contributes by removing each one from the entire model. Below ``ALL'' denotes using all components of GCAN. By removing dual co-attention, GRU-based representation, graph-aware representation, and CNN-based representation, we have sub-models ``-A'', ``-R'', ``-G'', and ``-C'', respectively. Sub-model ``-S-A'' denotes the one without both source tweet embeddings and dual co-attention. The results are presented in Figure~\ref{fig:ablation}. We can find every component indeed plays a significant contribution, especially for dual co-attention (``-A'') and the representation learning of user propagation and interactions (``-R'' and ``-G''). Since the source tweet provides fundamental clues, the accuracy drops significantly without it (``-S-A''). 

\subsection{GCAN Explainability}
The co-attention weights derived from Section~\ref{sec:dcm} attended on source tweet words and retweet users (source-propagation co-attention) allow our GCAN to be capable of explainability. By exhibiting where attention weights distribute, evidential words and users in predicting fake news can be revealed. Note that we do not consider source-interaction co-attention for explainability because user interaction features learned from the constructed graph cannot be intuitively interpretable.

\begin{figure}[!t]
\centering
\includegraphics[width=1.0\linewidth]{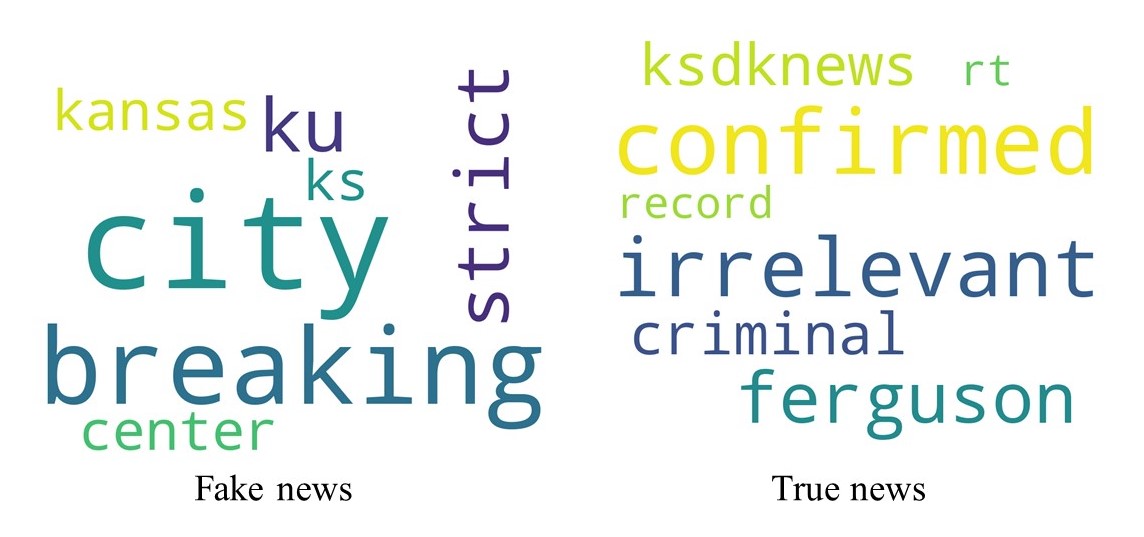}
\caption{Highlighting evidential words via word cloud. Larger font sizes indicate higher co-attention weights.}
\label{fig:cloud}
\end{figure}

\textbf{Explainability on Source Words.} To demonstrate the explainability, we select two source tweets in the test data. One is \textbf{fake} (``\textsl{breaking: ks patient at risk for ebola: in strict isolation at ku med center in kansas city \#kwch12}''), and the other is \textbf{real} (``\textsl{confirmed: this is irrelevant. rt @ksdknews: confirmed: \#mike-brown had no criminal record. \#ferguson}''). We highlight evidential words with higher co-attention weights in font sizes of word clouds, as exhibited in Figure~\ref{fig:cloud}. GCAN predicts the former to be fake with stronger attention on words ``breaking'' and ``strict'', and detects the latter as real since it contains ``confirmed'' and ``irrelevant.'' Such results may correspond to the common knowledge~\citep{fakel17a,fakel17b} that fake news tends to use dramatic and obscure words while real news is attended by confirmed and fact checking-related words. 

\begin{figure}[!t]
\centering
\includegraphics[width=1.0\linewidth]{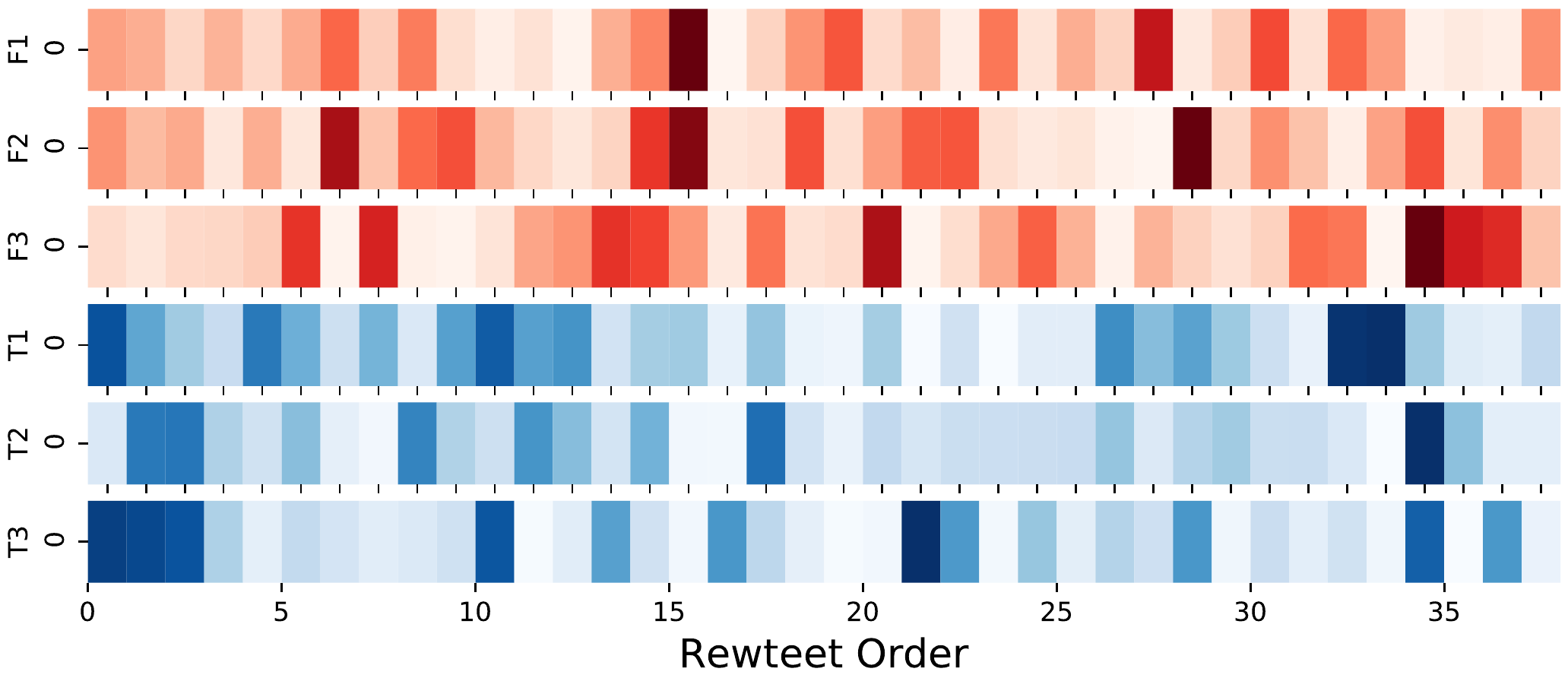}
\caption{Visualization of attention weights for user propagations of 3 fake (upper F1-F3) and 3 true source tweets. From left to right is retweet order. Dark colors refer to higher attention weights. }\label{fig:prop}
\end{figure}

% \begin{figure}[!t]
% \centering
% \includegraphics[width=0.75\linewidth]{frequency.pdf}
% \caption{Cumulated high attended retweet order 1-17.}
% \label{fig:freq}
% \end{figure}

\begin{figure}[!t]
\centering
\includegraphics[width=1.0\linewidth]{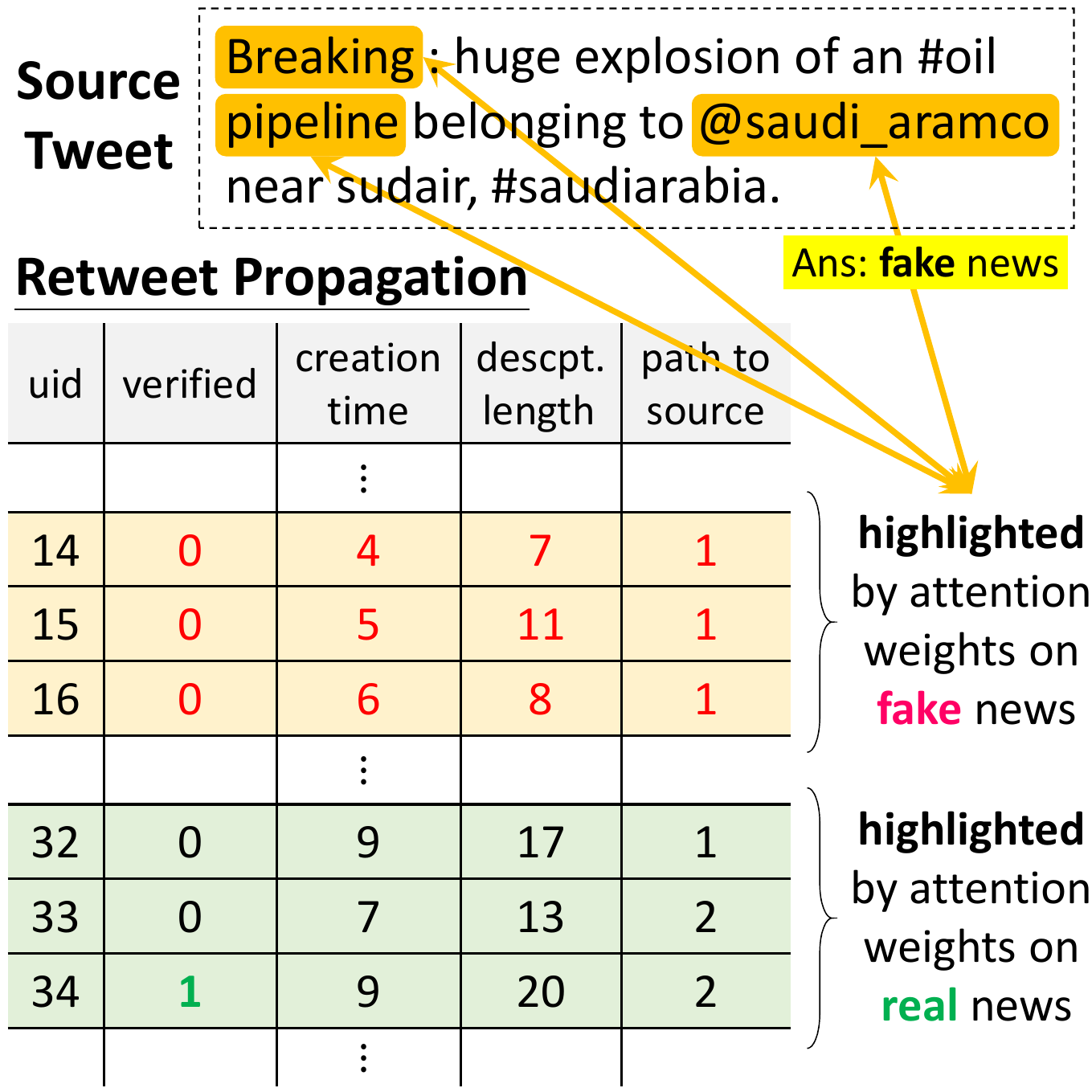}
\caption{Evidential words highlighed by GCAN in source tweet (upper) and suspicious users highlighed by GCAN in retweet propagation (bottom), in which each column is a user characteristic. Note that only few user characteristics are presented.}
\label{fig:case}
\end{figure}

\textbf{Explainability on Retweet Propagation.}
We aim to exploit the retweet order in propagations to unfold the behavior difference between fake and real news. We randomly pick three fake (F1-F3) and three true (T1-T3) source stories, and plot their weights from source-propagation co-attention (Section~\ref{sec:dcm}), as exhibited in Figure~\ref{fig:prop}, in which the horizontal direction from left to right denotes the order of retweet. The results show that to determine whether a story is fake, one should first examine the characteristics of users who \textbf{early} retweet the source story. The evidences of fake news in terms of user characteristics may be evenly distributed in the propagation.

\textbf{Explainability on Retweeter Characteristics.}
The source-propagation co-attention of our GCAN model can further provide an explanation to unveil the traits of suspicious users and the words they focus on. A case study is presented in Figure~\ref{fig:case}. We can find that the traits of suspicious users in retweet propagation can be: accounts are not verified, shorter account creation time, shorter user description length, and shorter graph path length to the user who posts the source tweet. In addition, what they highly attend are words ``breaking'' and ``pipeline.'' We think such kind of explanation can benefit interpret the detection of fake news so as to understand their potential stances.

\section{Conclusion}
\label{sec:conclude}
In this study, we propose a novel fake news detection method, Graph-aware Co-Attention Networks (GCAN). GCAN is able to predict whether a short-text tweet is fake, given the sequence of its retweeters. The problem scenario is more realistic and challenging than existing studies. Evaluation results show the powerful effectiveness and the reasonable explainability of GCAN. Besides, GCAN can also provide early detection of fake news with satisfying performance. We believe GCAN can be used for not only fake news detection, but also other short-text classification tasks on social media, such as sentiment detection, hate speech detection, and tweet popularity prediction. We will explore model generalization in the future work. Besides, while fake news usually targets at some events, we will also extend GCAN to study how to remove event-specific features to further boost the performance and explainability.

\section*{Acknowledgments}
This work is supported by Ministry of Science and Technology (MOST) of Taiwan under grants 109-2636-E-006-017 (MOST Young Scholar Fellowship) and 108-2218-E-006-036, and also by Academia Sinica under grant AS-TP-107-M05.

% The acknowledgments should go immediately before the references. Do not number the acknowledgments section.
% Do not include this section when submitting your paper for review.

\bibliography{acl2020}

\begin{thebibliography}{33}
\expandafter\ifx\csname natexlab\endcsname\relax\def\natexlab#1{#1}\fi

\bibitem[{Allcott and Gentzkow(2017)}]{fake17sv2}
Hunt Allcott and Matthew Gentzkow. 2017.
\newblock Social media and fake news in the 2016 election.
\newblock \emph{The Journal of Economic Perspectives}, 31:211--235.

\bibitem[{Castillo et~al.(2011)Castillo, Mendoza, and Poblete}]{textf11}
Carlos Castillo, Marcelo Mendoza, and Barbara Poblete. 2011.
\newblock Information credibility on twitter.
\newblock In \emph{Proceedings of the 20th International Conference on World
  Wide Web}, WWW '11, pages 675--684.

\bibitem[{Cha et~al.(2020)Cha, Gao, and Li}]{cacm20}
Meeyoung Cha, Wei Gao, and Cheng-Te Li. 2020.
\newblock Detecting fake news in social media: An asia-pacific perspective.
\newblock \emph{Commun. ACM}, 63(4):68--71.

\bibitem[{Chung et~al.(2014)Chung, Gulcehre, Cho, and Bengio}]{chung14e}
Junyoung Chung, Caglar Gulcehre, KyungHyun Cho, and Yoshua Bengio. 2014.
\newblock Empirical evaluation of gated recurrent neural networks on sequence
  modeling.

\bibitem[{Feng et~al.(2019)Feng, Hsu, Li, Yeh, and Lin}]{marine19}
Ming-Han Feng, Chin-Chi Hsu, Cheng-Te Li, Mi-Yen Yeh, and Shou-De Lin. 2019.
\newblock Marine: Multi-relational network embeddings with relational proximity
  and node attributes.
\newblock In \emph{The World Wide Web Conference}, WWW '19, pages 470--479.

\bibitem[{Guo et~al.(2019)Guo, Cao, Zhang, Shu, and Yu}]{emo19}
Chuan Guo, Juan Cao, Xueyao Zhang, Kai Shu, and Miao Yu. 2019.
\newblock Exploiting emotions for fake news detection on social media.
\newblock \emph{CoRR}, abs/1903.01728.

\bibitem[{Horne and Adali(2017)}]{fakel17b}
Benjamin Horne and Sibel Adali. 2017.
\newblock This just in: Fake news packs a lot in title, uses simpler,
  repetitive content in text body, more similar to satire than real news.
\newblock In \emph{Proceedings of AAAI International Conference on Web and
  Social Media}, pages 759--766.

\bibitem[{Jiang et~al.(2018)Jiang, Li, Chen, and Wang}]{shared18}
Jyun-Yu Jiang, Cheng-Te Li, Yian Chen, and Wei Wang. 2018.
\newblock Identifying users behind shared accounts in online streaming
  services.
\newblock In \emph{The 41st International ACM SIGIR Conference on Research \&
  Development in Information Retrieval}, SIGIR '18, pages 65--74.

\bibitem[{Kipf and Welling(2017)}]{gcn16}
Thomas~N. Kipf and Max Welling. 2017.
\newblock {Semi-Supervised Classification with Graph Convolutional Networks}.
\newblock In \emph{Proceedings of the 5th International Conference on Learning
  Representations}, ICLR '17.

\bibitem[{Kwak et~al.(2010)Kwak, Lee, Park, and Moon}]{twitter10}
Haewoon Kwak, Changhyun Lee, Hosung Park, and Sue Moon. 2010.
\newblock What is twitter, a social network or a news media?
\newblock In \emph{Proceedings of the 19th International Conference on World
  Wide Web}, WWW '10, pages 591--600.

\bibitem[{Kwon et~al.(2017)Kwon, Cha, and Jung}]{rfc17}
Sejeong Kwon, Meeyoung Cha, and Kyomin Jung. 2017.
\newblock Rumor detection over varying time windows.
\newblock \emph{PLOS ONE}, 12(1):1--19.

\bibitem[{Li et~al.(2018)Li, Lin, and Yeh}]{fpid18}
Cheng-Te Li, Yu-Jen Lin, and Mi-Yen Yeh. 2018.
\newblock Forecasting participants of information diffusion on social networks
  with its applications.
\newblock \emph{Information Sciences}, 422:432 -- 446.

\bibitem[{Liu and Wu(2018)}]{crnn18p}
Yang Liu and Yi-Fang Wu. 2018.
\newblock Early detection of fake news on social media through propagation path
  classification with recurrent and convolutional networks.
\newblock In \emph{AAAI Conference on Artificial Intelligence}, pages 254--261.

\bibitem[{Lu et~al.(2016)Lu, Yang, Batra, and Parikh}]{coatt16}
Jiasen Lu, Jianwei Yang, Dhruv Batra, and Devi Parikh. 2016.
\newblock Hierarchical question-image co-attention for visual question
  answering.
\newblock In \emph{Proceedings of the 30th International Conference on Neural
  Information Processing Systems}, NIPS'16, pages 289--297.

\bibitem[{Ma et~al.(2016)Ma, Gao, Mitra, Kwon, Jansen, Wong, and Cha}]{rnn16}
Jing Ma, Wei Gao, Prasenjit Mitra, Sejeong Kwon, {Bernard J.} Jansen, {Kam Fai}
  Wong, and Meeyoung Cha. 2016.
\newblock Detecting rumors from microblogs with recurrent neural networks.
\newblock \emph{IJCAI International Joint Conference on Artificial
  Intelligence}, pages 3818--3824.

\bibitem[{Ma et~al.(2015)Ma, Gao, Wei, Lu, and Wong}]{svmts15}
Jing Ma, Wei Gao, Zhongyu Wei, Yueming Lu, and Kam-Fai Wong. 2015.
\newblock Detect rumors using time series of social context information on
  microblogging websites.
\newblock In \emph{Proceedings of the 24th ACM International on Conference on
  Information and Knowledge Management}, CIKM '15, pages 1751--1754.

\bibitem[{Ma et~al.(2017)Ma, Gao, and Wong}]{kerl17s}
Jing Ma, Wei Gao, and {Kam Fai} Wong. 2017.
\newblock Detect rumors in microblog posts using propagation structure via
  kernel learning.
\newblock In \emph{ACL 2017 - 55th Annual Meeting of the Association for
  Computational Linguistics, Proceedings of the Conference}, pages 708--717.

\bibitem[{Ma et~al.(2018)Ma, Gao, and Wong}]{trnn18}
Jing Ma, Wei Gao, and Kam-Fai Wong. 2018.
\newblock Rumor detection on twitter with tree-structured recursive neural
  networks.
\newblock In \emph{Proceedings of the 56th Annual Meeting of the Association
  for Computational Linguistics}, pages 1980--1989.

\bibitem[{Popat(2017)}]{lang17}
Kashyap Popat. 2017.
\newblock Assessing the credibility of claims on the web.
\newblock In \emph{Proceedings of the 26th International Conference on World
  Wide Web Companion}, WWW '17 Companion, pages 735--739.

\bibitem[{Potthast et~al.(2018)Potthast, Kiesel, Reinartz, Bevendorff, and
  Stein}]{wstyle18}
Martin Potthast, Johannes Kiesel, Kevin Reinartz, Janek Bevendorff, and Benno
  Stein. 2018.
\newblock A stylometric inquiry into hyperpartisan and fake news.
\newblock In \emph{Proceedings of the 56th Annual Meeting of the Association
  for Computational Linguistics}, ACL '18, pages 231--240.

\bibitem[{Rashkin et~al.(2017)Rashkin, Choi, Jang, Volkova, and
  Choi}]{fakel17a}
Hannah Rashkin, Eunsol Choi, Jin~Yea Jang, Svitlana Volkova, and Yejin Choi.
  2017.
\newblock Truth of varying shades: Analyzing language in fake news and
  political fact-checking.
\newblock In \emph{Proceedings of the 2017 Conference on Empirical Methods in
  Natural Language Processing}, pages 2931--2937.

\bibitem[{Reis et~al.(2019)Reis, Correia, Murai, Veloso, and
  Benevenuto}]{fexp19}
Julio C.~S. Reis, Andr{\'e} Correia, Fabr\'{\i}cio Murai, Adriano Veloso, and
  Fabr\'{\i}cio Benevenuto. 2019.
\newblock Explainable machine learning for fake news detection.
\newblock In \emph{Proceedings of the 10th ACM Conference on Web Science},
  WebSci '19, pages 17--26.

\bibitem[{Ruchansky et~al.(2017)Ruchansky, Seo, and Liu}]{csi17s}
Natali Ruchansky, Sungyong Seo, and Yan Liu. 2017.
\newblock Csi: A hybrid deep model for fake news detection.
\newblock In \emph{Proceedings of the 2017 ACM on Conference on Information and
  Knowledge Management}, CIKM '17, pages 797--806.

\bibitem[{Sampson et~al.(2016)Sampson, Morstatter, Wu, and Liu}]{lis16s}
Justin Sampson, Fred Morstatter, Liang Wu, and Huan Liu. 2016.
\newblock Leveraging the implicit structure within social media for emergent
  rumor detection.
\newblock In \emph{Proceedings of the 25th ACM International on Conference on
  Information and Knowledge Management}, CIKM '16, pages 2377--2382.

\bibitem[{Shu et~al.(2019{\natexlab{a}})Shu, Cui, Wang, Lee, and Liu}]{kshu19h}
Kai Shu, Limeng Cui, Suhang Wang, Dongwon Lee, and Huan Liu.
  2019{\natexlab{a}}.
\newblock defend: Explainable fake news detection.
\newblock In \emph{Proceedings of the 25th ACM SIGKDD International Conference
  on Knowledge Discovery \& Data Mining}, KDD '19, pages 395--405.

\bibitem[{Shu et~al.(2017)Shu, Sliva, Wang, Tang, and Liu}]{fake17sv1}
Kai Shu, Amy Sliva, Suhang Wang, Jiliang Tang, and Huan Liu. 2017.
\newblock Fake news detection on social media: A data mining perspective.
\newblock \emph{SIGKDD Explor. Newsl.}, 19(1):22--36.

\bibitem[{Shu et~al.(2019{\natexlab{b}})Shu, Zhou, Wang, Zafarani, and
  Liu}]{kshu19p}
Kai Shu, Xinyi Zhou, Suhang Wang, Reza Zafarani, and Huan Liu.
  2019{\natexlab{b}}.
\newblock The role of user profile for fake news detection.
\newblock \emph{CoRR}, abs/1904.13355.

\bibitem[{Wang and Li(2019)}]{osne19}
Pei-Chi Wang and Cheng-Te Li. 2019.
\newblock Spotting terrorists by learning behavior-aware heterogeneous network
  embedding.
\newblock In \emph{Proceedings of the 28th ACM International Conference on
  Information and Knowledge Management}, CIKM '19, pages 2097--2100.

\bibitem[{Wang et~al.(2018)Wang, Ma, Jin, Yuan, Xun, Jha, Su, and
  Gao}]{eann18h}
Yaqing Wang, Fenglong Ma, Zhiwei Jin, Ye~Yuan, Guangxu Xun, Kishlay Jha, Lu~Su,
  and Jing Gao. 2018.
\newblock Eann: Event adversarial neural networks for multi-modal fake news
  detection.
\newblock In \emph{Proceedings of the 24th ACM SIGKDD International Conference
  on Knowledge Discovery \&\#38; Data Mining}, KDD '18, pages 849--857.

\bibitem[{Yan et~al.(2015)Yan, Yen, Li, Zhao, and Hu}]{tshort15}
Rui Yan, Ian~E.H. Yen, Cheng-Te Li, Shiqi Zhao, and Xiaohua Hu. 2015.
\newblock Tackling the achilles heel of social networks: Influence propagation
  based language model smoothing.
\newblock In \emph{Proceedings of the 24th International Conference on World
  Wide Web}, WWW '15, pages 1318--1328.

\bibitem[{Yang et~al.(2012)Yang, Liu, Yu, and Yang}]{yangp12}
Fan Yang, Yang Liu, Xiaohui Yu, and Min Yang. 2012.
\newblock Automatic detection of rumor on sina weibo.
\newblock In \emph{Proceedings of the ACM SIGKDD Workshop on Mining Data
  Semantics}, MDS '12.

\bibitem[{Yang et~al.(2018)Yang, Zheng, Zhang, Cui, Li, and Yu}]{tcnn18}
Yang Yang, Lei Zheng, Jiawei Zhang, Qingcai Cui, Zhoujun Li, and Philip~S. Yu.
  2018.
\newblock Ti-cnn: Convolutional neural networks for fake news detection.

\bibitem[{Zhao et~al.(2015)Zhao, Resnick, and Mei}]{enquiry15}
Zhe Zhao, Paul Resnick, and Qiaozhu Mei. 2015.
\newblock Enquiring minds: Early detection of rumors in social media from
  enquiry posts.
\newblock In \emph{Proceedings of the 24th International Conference on World
  Wide Web}, WWW '15, pages 1395--1405.

\end{thebibliography}
\bibliographystyle{acl_natbib}

\end{document}